\documentclass[letterpaper, 10 pt, conference]{ieeeconf}  

\IEEEoverridecommandlockouts                              

\usepackage{graphics} 
\usepackage{epsfig} 
\usepackage{multirow}
\usepackage{tikz}
\usetikzlibrary{arrows}
\tikzstyle{int}=[draw, fill=blue!20, minimum size=2em]
\tikzstyle{init} = [pin edge={to-,thin,black}]
\usepackage[latin1]{inputenc}
\usepackage[T1]{fontenc}
\usepackage{url}

\usepackage{booktabs}

\usepackage{caption}
\DeclareCaptionFont{black}{ \color{black} }
\DeclareCaptionFormat{listing}{
	\parbox{\columnwidth}{\hspace{15pt}#1#2#3}
}
\usepackage[textwidth=\marginparwidth,figwidth=\columnwidth]{todonotes} 
\presetkeys{todonotes}{size=\scriptsize, color=green!30}{} 

\usepackage{paralist}

\title{\LARGE \bf
	Towards Robot-independent Manipulation Behavior Description
}

\author{Malte Wirkus$^{1}$
	\thanks{$^{1}$Malte Wirkus is with Deutsches Forschungszentrum für Künstliche Intelligenz, Robotics Innovation Center,
		Bremen, Germany
		{\tt\small malte.wirkus@dfki.de}}%
}

\usepackage{listings}
\usepackage{color}

\definecolor{dkgreen}{rgb}{0,0.6,0}
\definecolor{gray}{rgb}{0.5,0.5,0.5}
\definecolor{mauve}{rgb}{0.58,0,0.82}

\usepackage{caption}[2013/01/01] 
\DeclareCaptionFormat{listing}
{
	{\parbox{\dimexpr\columnwidth-2\fboxsep}{\centering#1\kern-.35em#2\kern-.35em#3}}}
\renewcommand{\vec}{\textbf}

\newcommand{\varsystate}{\ensuremath{\vec x}}
\newcommand{\vartime}{\ensuremath{t}}
\newcommand{\varsysstateattime}{\ensuremath{\vec x_\vartime}}
\newcommand{\varmotionparams}{\ensuremath{\Theta}}
\newcommand{\varsysrefattime}{\ensuremath{\hat\varsysstateattime}}

\begin{document}

	\maketitle
	\thispagestyle{empty}
	\pagestyle{empty}

	\begin{abstract}
		In this paper we present a workflow to design and control robot manipulation behavior. To remain independent from particular robot hardware and an explicit area of application, an embedded domain specific language (eDSL) is used to describe the particular robot and a controller network that drives the robot. We make use of 
		\begin{inparaenum}[\itshape a\upshape)]
			\item a component-based software framework,
			\item model-based algorithms for motion- and sensor processing representations,
			\item an abstract model of the control system, and
			\item a plan management software,
		\end{inparaenum}
		to describe a sequence of software component networks that generate the desired robot behavior.
		
		As first results, we present an eDSL for the description of a robotic system composed of mechatronic subsystems, and for the creation of a multi-stage control network.
	\end{abstract}

	\section{INTRODUCTION}
	Software frameworks for the development of robotic systems such as ROS \cite{ROS2009} or Rock \cite{Joyeux2014}, provide tools to support software development and also define a common component interface for the software created within that framework. This has a significant impact on robot programming: While the tools increase the developer's productivity, reusability of software is increased by the common component interface. This leads to a situation where today's roboticist can benefit from a rich collection of software components. Increasingly, robot programming becomes an integration and configuration task.
	
	Using a component-based software architecture, the robot's behavior is defined by the selection of individual components, their configuration, and the component interconnections. The reuse of software components can save a lot of development work (mainly on the algorithmic side), but the integration process has its own challenges and still can be laborious \cite{Schwendner2014}. Abstract modeling of recurring subsystems, and the use of domain specific programming languages provide a promising tool to support building up operative systems from single components \cite{Joyeux2011}. In this paper, we introduce our approach for their utilization on describing robot behavior. Of special interest for us is the ability to describe manipulation behavior and apply this description to robots of different morphology and hardware.
	
	In the next section we will introduce a workflow which eventually should allow the generation of a broad range of robot manipulation behavior without explicitly programming any controllers. The workflow allows decoupling the description of the action to perform from the actual robot hardware by modeling both independently. Section \ref{sec:control_system} shows first results: An embedded domain specific language (eDSL) for the creation of controller hierarchies out of arbitrary robotic subsystems. Since this work is based on the Rock framework, the relevant aspects about it are explained beforehand, in Section \ref{sec:rock}.

	\section{MANIPULATION BEHAVIOR GENERATION}\label{sec:behavior_design}
	Within the scope of this work, behavior of a robot is defined as the motion the robot performs. The behavior is created by the controller that drive the robot's joint. The controllers are constructed from a processing chain where data perceived fromt he robot's sensors gets transformed into motion commands. Also, the data gets transformed from the sensor reference frame to the reference frame of the controlled manipulator, and finally to joint level commands. This leads to a situation, where the controller needs to be aware of the robots kinematics to do the required frame transformations, and is also aware of the motion to perform in order to accomplish the task at hand -- the controller is dependent on both, the task and the robot's morphology.
	
	\begin{figure}[tb]
		\centering
		\includegraphics[width=0.47\textwidth]{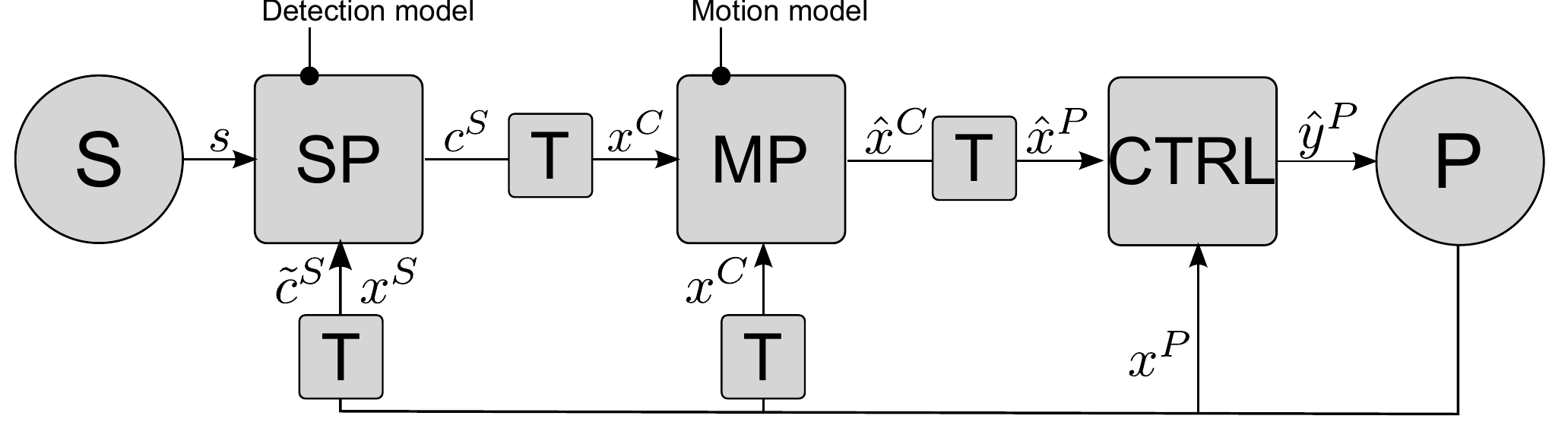}
		\caption{A sensor processing pipeline (SP) extracts a particular object (context, $C$) from a data stream of a sensor $S$. A motion model is executed, which encodes the motion plan necessary to perform a specific action. The output is used as set point for a controller that drives a robotic system. Each step requires its data expressed in different reference frames. A transformation module takes care of providing the data in the proper reference frames.}\label{fig:basic_processing}
	\end{figure}
	
	We suggest to decouple the the robot's morphology from the task description by dividing a controller into separate parts, where each part is expressed in a suitable reference frame (cf. Fig. \ref{fig:basic_processing}):
	\begin{inparaenum}[\itshape 1\upshape)] 
		\item Through sensors the robot captures its environment. This data gets processed by a sensor processing pipeline (SP), that ultimately transforms the raw sensor data into a controllable quantity, i.e. positions in Cartesian space $c^S$ expressed in the sensor's reference frame.
		\item A motion generation part where an action specific motion plan (MP) is executed and transformed into motion commands $\hat x^C$. The motion plan describes the motion of the robot's end effector, to fulfill a specific action relative to the contex of the action (e.g. the object to manipulate). If this motion is expressed in a reference frame representing the goal of the motion, the execution of the motion plan is independent from the robot's morphology. 
		\item The resulting data $\varsysrefattime$ is transformed into the reference frame of the robotic system (P) and serves as set point for a controller, that generates a control command $\hat y^P$ for the system. The system on the other hand provides an observable variable $x^P$, which, when transformed into the respective reference frames, can be used as feedback in each step of processing.
	\end{inparaenum}
	
	With the term \emph{motion plan} we refer to a function that represents a particular movement, independent of the underlying technique used. A motion plan can for example be a function $\varsysrefattime = f(\varmotionparams, \varsystate)$, relating a set of algorithm specific motion parameters $\varmotionparams$ together with a representation of the current system state $\varsysstateattime$ to a new system state $\varsysrefattime$ that serves as reference. A waypoint navigation controller would fall into this category, as well as an attractor field. Motion plans could also be time dependent $\varsysrefattime = f(\varmotionparams, \vartime)$ or both, time- and state dependent $\varsysrefattime = f(\varmotionparams, \vartime, \varsysstateattime)$ such as Dynamic Motion Primitives (DMP) \cite{Schaal2006}.
	
	Like the imitation learning community, we also believe that a broad range of behavior can be generated if parametric models are utilized \cite{Ijspeert03}\cite{Metzen2013} for encoding motion. Moreover, instead of providing a fixed architecture which only lets the motion model parameters open for influencing robot behavior, we want to suggest a complete robot application development workflow that allows to: 
	\begin{inparaenum}[\itshape 1\upshape)]
		\item Describe your robot as a composition of mechatronic parts.
		\item Program robot behavior by creating arbitrary subsystems and control them utilizing parametric motion descriptions while taking into account sensor information.
		\item Arrange different robot behaviors chronologically to allow completion of complex tasks.
		\item Potentially allow the transfer of behavior between robots.
	\end{inparaenum}
	The next subsection will describe this workflow in more detail.

	\begin{figure}[tbcp]
		\centering
		\includegraphics[width=\columnwidth]{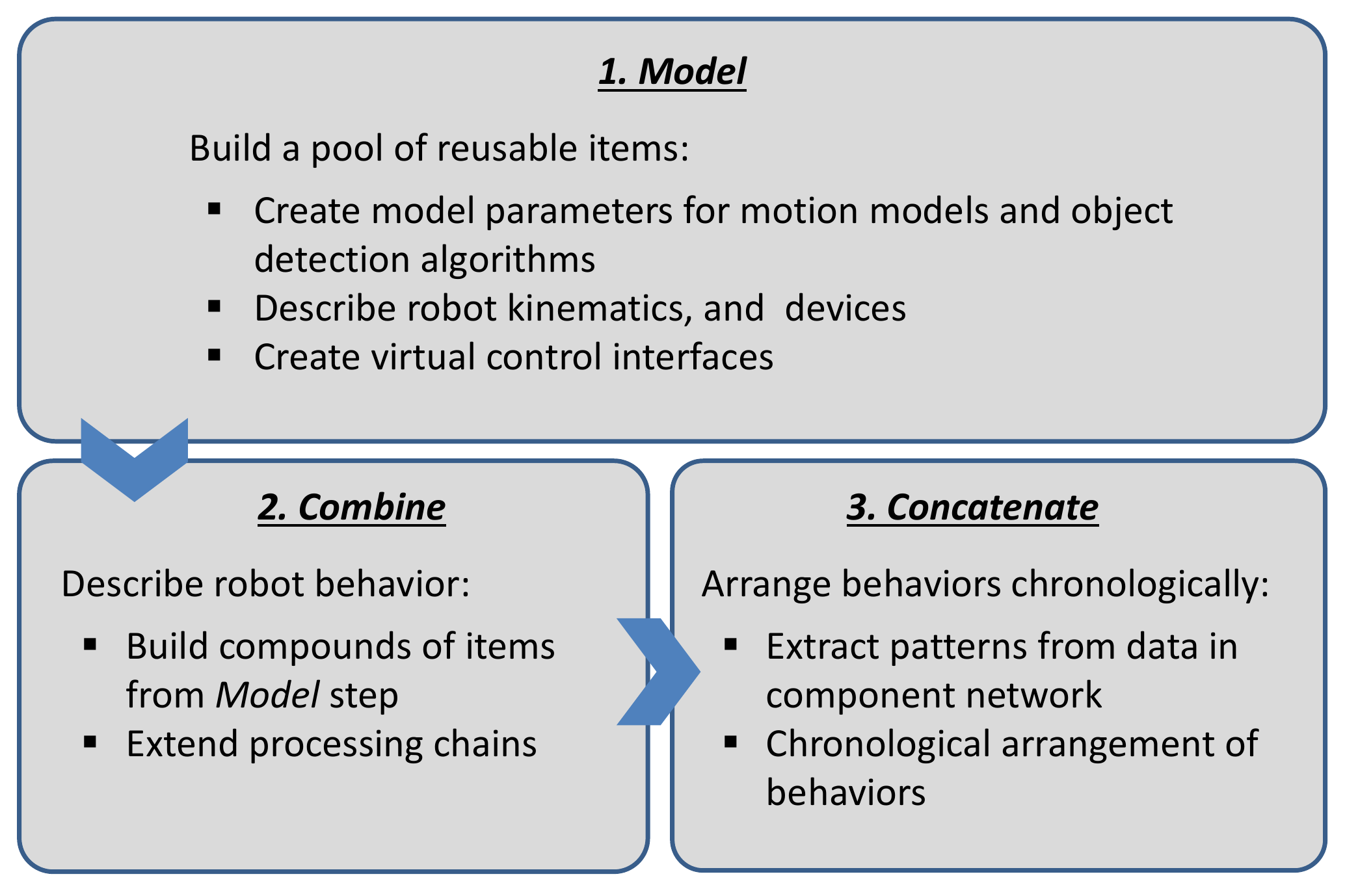}
		\caption{The proposed workflow is separated into three steps:  \emph{Model}, \emph{Connect} and \emph{Concatenate}.}\label{fig:anwendungsentwicklung}
	\end{figure}
	\subsection{Application Development Workflow}
	
	The application development process in the proposed workflow is outlined in Fig. \ref{fig:anwendungsentwicklung}. It is separated into the three steps \emph{Model}, \emph{Combine} and \emph{Concatenate}. Each of these steps will be described individually in the following paragraphs.
	
	In the first step (\emph{Model}) a set of models is created. This includes algorithm-specific models for motion representation and object detection. For their application, object detection models are dependent on the sensor type used and the existence of a detector component able to interpret the model. Apart from a compatible sensor, the detection models are independent from particular robot hardware and thus could be shared between different systems. The motion representations are also dependent on the existence of an interpreting algorithm. Their formulation however could, as mentioned earlier, be done in a robot-agnostic fashion. Finally, the actual robot needs to be modeled by describing its kinematics and devices (sensors and joints). Devices are controlled with their device drivers which are implemented as software components. Within the kinematic description, positions of joints and structural segments, but also positions of sensors are included. To create a link between software representation of a device and the embodiment of the robot, devices drivers are assigned to respective elements within the kinematic description of the robot. Furthermore, functional units are defined from the robot's joints (\emph{virtual control system}). We go more into detail about this subject in Section \ref{sec:control_system}.
	
	In the second step (\emph{Combine}), these elements are related to each other to build compounds of motion models, virtual control systems, sensor processing models and sensors. Additionally sensor processing chains or control chains can be extended by inserting additional data processors from an extendable pool of categorized software components (see Section \ref{sec:control_system} for details). These compounds implicitly describe the software component network that generates the desired behavior. To support the application developer, the tooling should check the compounds for plausibility.
	
	On the base of these two steps, a broad range of different system behaviors can be described. For practical application, different behaviors are required to be coordinated over a longer time scale. This coordination is described in the third application development step called \emph{Concatenate}. Component network descriptions are arranged chronologically. During runtime, data on the currently running component network's ports represent the current system state. With mapping patterns within this data to switching of behavior networks, such coordination could be achieved.
	
	In this section, we described a general workflow to express robot manipulation actions independent of a specific target system, and also described necessary steps to assign these actions to an actual robotic system and an explicit context. To make this a feasible process, tools need to be developed that support the application developer in each steps of the workflow. In our implementation, we rely on the Rock framework, whose relevant aspects will be introduced briefly in the next section.

	\section{Component Networks and Plan Management}\label{sec:rock}
	The Robot Construction Kit (Rock) \cite{Joyeux2014} is a framework to develop software for robotic systems. Its component model is based on the Orocos Real Time Toolkit \cite{Soetens2005}, and additionally provides a set of tools for development of software for robots.
	
	\subsection{Component models}
	Components within Rock are defined by a data flow and a configuration interface as well as a common life-cycle. The life-cycle reflects the transitions of states a component can go through. These states are inoperative states for configuration or stopped components, runtime, and error states. It is important to mention that different modes of operation of an algorithm should not be expressed through different runtime states, but through different components. The rationale behind this is, that a single component should only have a single well-defined operative mode, and not switch its behavior in different runtime-modes. This makes the behavior of a component solely dependent on its configuration and the input data, but not on its history -- a property that increases the possibility of re-using single components and integrating them into higher control levels. It also allows to label components with simple semantic labels that represent the kind of processing they implement \cite{Joyeux2014}.
	
	\subsection{Plan management}
	The plan manger Roby \cite{Joyeux2010} allows representing, executing and also adapting plans. An activity within a plan is defined by a graph of task relations. Task in this context means an operation of the system which is implemented by a block of user code. The relations between the tasks represent semantic dependencies between tasks, which are used to express why a task is within the plan. Progress of tasks is modeled using events representing an identifiable situation that occurred during task execution. Events appear in two types: \emph{controllable} and \emph{contingent}. Controllable events can be triggered by the plan manager itself by calling a procedure on the task that deterministically brings the task in the situation that emits the event. Contingent events are non-controllable and thus are triggered by other processes but the plan manager, i.e. usually the task's operation. Long-term system behavior can be created by linking emitted events to the execution of controllable events.
	
	\subsection{Modeling component network models}
	Abstract descriptions of component networks can be created with the tool Syskit \cite{Joyeux2011}. Syskit provides a Ruby based eDSL to describe compositions of
	\begin{inparaenum}[\itshape a\upshape)]
		\item actually implemented software components,
		\item an abstract operation, or
		\item another component network already modeled with Syskit,
	\end{inparaenum}
	in order to model abstract functional component networks. The abstract operations (called \emph{data services}) are defined by a minimal data flow interface a component must provide and a semantic label representing their functionality. To construct a fully functional component network from an abstract one, an instantiation specification must be created by selecting actual components or subnetworks as replacement for the abstract data services. During the instantiation of a component network description, redundancies are merged, which might occur from using identical components in multiple subnetworks or from already running components. This allows the transition of an arbitrary running component network to a new one, what leads to the ability to embed different component network descriptions into a higher-level coordination. Syskit was integrated with Roby such that Syskit's compositions can be used as Roby tasks.
	
	The tools described above already provide a great starting point in the realization of the workflow presented in Section \ref{sec:behavior_design}. Rock's component model is used to implement software components, but also a large pool of already implemented algorithms is available within the Rock's infrastructure. Syskit is used to semantically group and model relations between those components. Roby plans are used as representation for sequences of different behaviors. In the following section, we want to introduce a new language layer implemented as extension to Syskit's eDSL, which uses Ruby as host language. It allows the efficient description of a multi-stage control system.

	\section{CONTROL SYSTEM SPECIFICATION}\label{sec:control_system}
	Complex robots are often composed from multiple mechatronic subsystems. Each of these subsystems is connected with its own interface to the control computer. Each subsystem can be accessed individually with their respective drivers, and is fully functional on its own without the other subsystems.
	
	However, for the robot application development, different subsystems might be required in order to perform a specific task. We enable the specification of multiple \emph{virtual control systems} by providing an eDSL that features
	\begin{inparaenum}[\itshape a\upshape)]
		\item the description of a robot as aggregation of mechatronic subsystems,
		\item free recombination of elements of these subsystems, and
		\item assignment of cascaded controllers to these virtual control interfaces.
	\end{inparaenum}
	The interpretation of the program results in the configuration of a set of \emph{core components} (see Table \ref{tab:core-component}), as well as the creation of the instantiation specification for a Syskit composition representing the control network.
	
	\newcommand{\pcell}[2][c]{%
		\begin{tabular}[#1]{@{}l@{}}#2\end{tabular}}
	
	\begin{table}[tbp]
		\definecolor{shadecolor}{rgb}{0.85,0.85,0.95}
		\scriptsize
		\centering
		\begin{tabular}{p{0.13\columnwidth}p{0.77\columnwidth}}
			\toprule
			\textbf{Dispatch}: & From a stream of joint data, allows to extract any subset of joints and write it on a dedicated port. \\[.5ex]
			\textbf{Kinematics}: & Calculates forward/inverse kinematics for any joint chain within the kinematic description of the robot. \\[.5ex]
			\textbf{Controllers}: & General purpose controller components (e.g. PID) operating in Cartesian and joint space.\\[.5ex]
			\textbf{Transformer}: & Allows the calculation of arbitrary transforms if they can be determined from a given graph of known frame transforms. \\[.5ex]
			\textbf{Wbc}: & Whole-body control. Generates a common control command for a set of constraints imposed on different parts of the robot \cite{Schutter2007}. \\[.5ex]
			\bottomrule
		\end{tabular} 
		
		\caption{Core components}\label{tab:core-component}
	\end{table}

	For the remainder of this section, we will present the description of a control network for a hypothetical robot with four degrees of freedom which are separated into two mechatronic subsystems. We will proceed bottom-up, starting with the device drivers.
	
	In the following code snippet (Listing \ref{lst:joint_driver}), we assume that a software component called \emph{MyJointDriver::Task} implementing the driver for the subsystems exists. We register this component to be a driver for a joint device, and also state that it can be operated in position and velocity control mode. The different control modes are then reflected by the joint device type.

	\begin{lstlisting}[caption=Joint driver registration,label=lst:joint_driver]
module Devices
	joints_device_type "MyJointsPositionDriver" do
		position_controlled
	end
	joints_device_type "MyJointsVelocityDriver" do
		velocity_controlled
	end
end

MyJointDriver::Task.driver_for Devices::MyJointsPositionDriver, :as => 'position_controlled'
MyJointDriver::Task.driver_for Devices::MyJointsVelocityDriver, :as => 'velocity_controlled'
	\end{lstlisting}
	
	We compose our robot as aggregation of the two subsystems (see Listing \ref{lst:robot}), both controlled with a driver of the previously defined type, but using different configurations. The mechanical structure of the robot is described in the file referred to in the \texttt{kinematic\_description} command, which is required for several core components to work. We use the Unified Robot Description Format (URDF)\footnote{URDF: \url{http://wiki.ros.org/urdf} (last accessed 08/30/14)} for describing the robot's kinematics.
	
	\begin{minipage}{\columnwidth}
		\begin{lstlisting}[caption=Robot definition,label=lst:robot]
robot do
	kinematic_description "/path/to/my/kinematic_description.urdf"
	device(Devices::JointsPositionDriver, :as => 'armr').joint_names('ar', 'br', 'cr').with_conf('armr')
	device(Devices::JointsPositionDriver, :as => 'arml').joint_names('al', 'bl', 'cl').with_conf('arml')
	device(Devices::JointsPositionDriver, :as => 'hr').joint_names('wr','gr').with_conf('hr')
	device(Devices::JointsPositionDriver, :as => 'hl').joint_names('wl','gl').with_conf('hl')
	device(Devices::JointsVelocityDriver, :as => 'head').joint_names('p', 't'). with_conf('head')
end	
		\end{lstlisting}
	\end{minipage}
	
	Each joint is uniquely identified by a name. The joints controlled by a joint device driver are specified by with the \texttt{joint\_names} function (cf. Listing \ref{lst:robot}). The names given here, must match the joint names used in the kinematic model, such that a link between the structural representation and the hardware driver specification is created. The joints described here are the resources we can use in the next step.
	
	In Listing \ref{lst:control_interfaces}, we show the creation of one stage in a multi-stage control network. Multiple virtual control interfaces can be created from the available joints inside a \texttt{control\_collection} block. We distinguish between the control modes \emph{position} and \emph{velocity} a subsystem can be operated. If the joint used for a virtual control interface provide a differnt control mode that the on requested, a suitable control command conversion component is interposed. We further distinguish between two methods to deal with controllers running in parallel: \emph{direct control} and \emph{whole-body control}. The former corresponds to the classical method, where a controller is attached to the controlled plant and directly modifies its control variable. This method does not allow the same joint being commanded by two different controllers at the same time, since this would result in unspecified behavior. The later method, \emph{whole-body control} allows controlling the same joints in parallel. This is made possible due to an algorithm that is capable to resolve conflicting control signals utilizing a hierarchical sorting of the controllers \cite{Schutter2007}. As it is shown in Listing \ref{lst:control_interfaces}, direct control interfaces can be defined right away using the \texttt{control\_interface} command. A set of whole-body control interfaces is defined as block given to the \texttt{wbc}-command, where each control interface is also assigned a priority and some algorithm specific attributes.
	A virtual control interface is represented by a port providing a data stream containing only status information of the corresponding joints, as well as an input port to receive commands for these joints on the \emph{joint dispatcher},  respectively the \emph{wbc} component (cf. Table \ref{tab:core-component}).
	
	\begin{lstlisting}[caption=Control interface definition,label=lst:control_interfaces]
control_collection "l2" do
	used_joints = ['ar','br','cr','wr','p','t']
	wbc_interface used_joints, :as => "wbc", :initial_joint_weights => [1]*used_joints.size do
		cartesian_control_interface ['O','WR'], 
			:as => "cart_arm_plus_wrist", 
			:joint_names => ['ar','br','cr','wr'], 
			:priority => 1, :weights => [1,1,1,0.5]
	
		control_interface ['p','t'], :as => "head", :priority => 2

		control_interface ['ar','br','cr','wr'], 
			:as => "body_posture", :priority => 3
	end
	
	control_interface ['gr'], 
		:control_mode => :position, 
		:as => 'finger'
	cartesian_control_interface 'O', 'WL', 
		:joint_names => ['al','bl','cl','wl'],  
		:control_mode => :velocity, 
		:as => 'other_arm'
	end
	
	cascade_control finger_interface do 
		push TrajectoryGeneration::Task
				.with_conf('arm_with_hand')
	end
end
	\end{lstlisting}

			\addtolength{\textheight}{-6.5cm}   
	
\begin{figure}[tbp]
	\centering
	\begin{tabular}{cc}
		{\includegraphics[width=0.45\columnwidth]{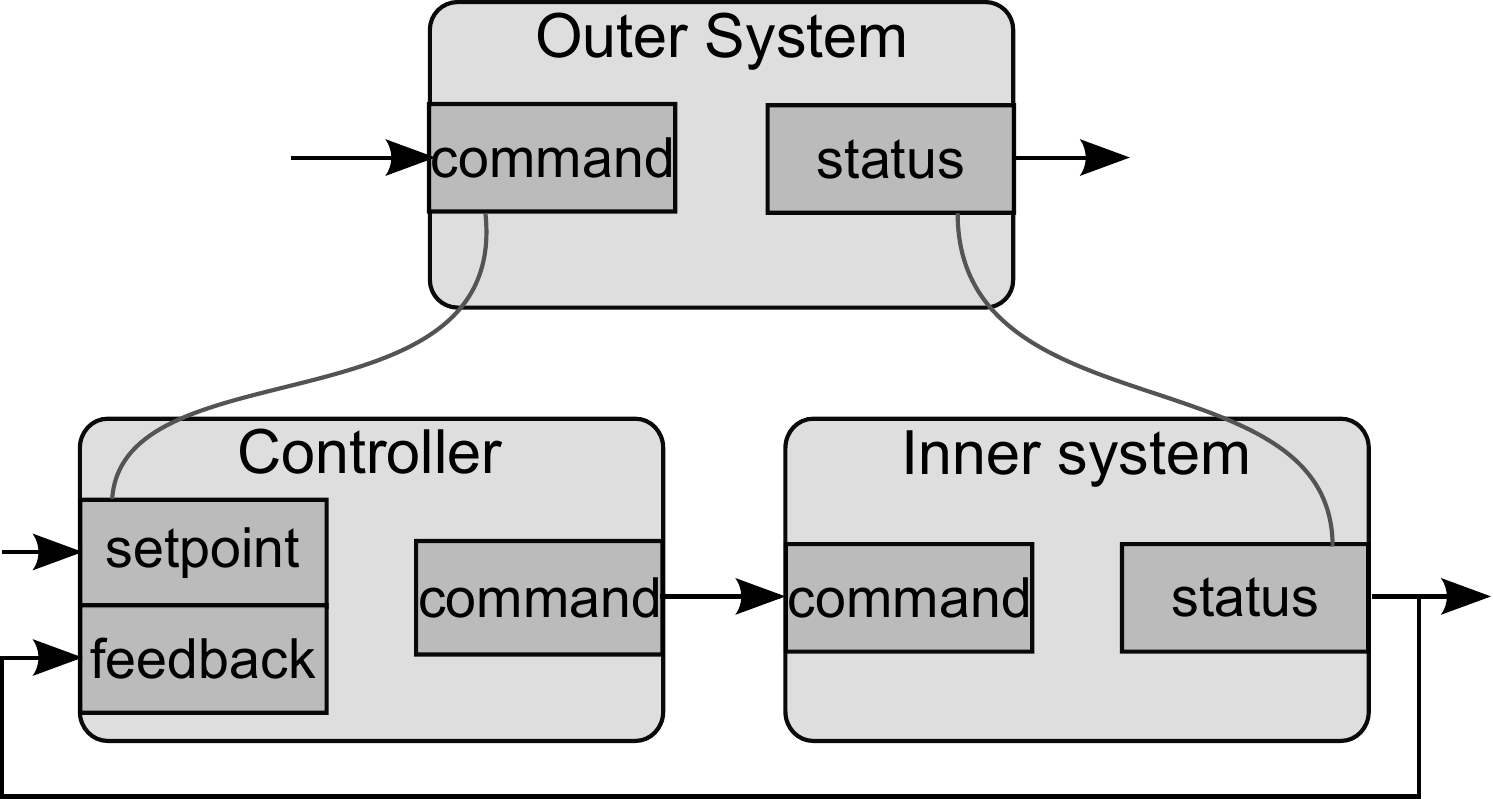}} & 
		{\includegraphics[width=0.45\columnwidth]{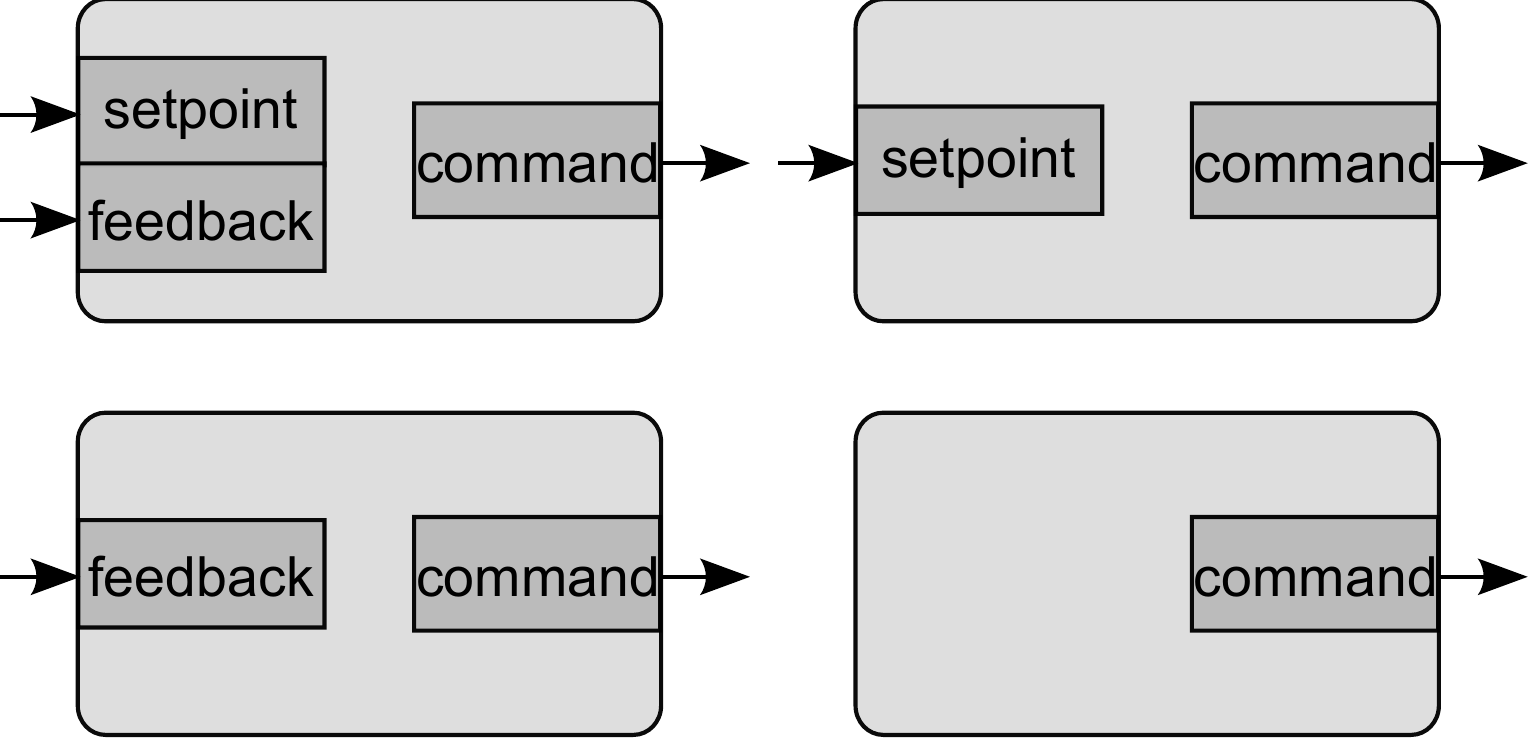}} \\
		(a) & (b)
	\end{tabular}
	\caption{(a) Model of a cascade controller. If the controller provides a port for set points, an \emph{outer control system} is created. (b) Dataflow interfaces that can be used in a cascade controller.}\label{fig:controller_cascade}
\end{figure}

	Also shown in the code snipped above, a cascade of controllers can be assigned to each control interface. Its definition is started with the \texttt{cascade\_control} command and the cascade is built by adding controllers with the \texttt{push} command. The command generated by the outer stage of the cascade is passed to the inner stage as set point signal (cf. Fig. \ref{fig:controller_cascade} (a)). The cascade controller allows a simple definition of data processing chains for control signals for example such that include interpolation, signal filtering or similar processes. A component qualifies for the use within a cascade controller if it fulfills one of the data services illustrated in Fig. \ref{fig:controller_cascade} (b). Whether further cascade stages can be preceded is determined by the existence of a set point port on the currently outer-most component.

	A processing task involving more complex interconnections could be modeled as a component network using Syskit compositions. If it fulfills one of the data services from  Fig. \ref{fig:controller_cascade} (b), it could also be used within a cascade controller.

	If the outer most controller of a cascade controller provides a set point port, it is exported from the controller collection. Multiple \texttt{control\_collections} can be created which are interconnected in a layered fashion. Hereby, the higher level control collection only has access to those joints which have been exported in the level below. 
	

	\section{CONCLUSIONS}
	We described a general workflow that decouples the description of robot manipulation behavior from the morphology of the robot and from explicit object representations. As first results, we presented an embedded domain specific language for the description of multi-stage control networks.
	
	Such a layered control system might be for instance useful to integrate a security layer, taking care of self-collision control or interpolation of all robot joints, below a layer that actually implements the task at hand using different virtual subsystems. Syskit's eDSL proved to be easily extendable to provide higher-level domain specific commands. Further development and research is necessary to accomplish the workflow proposed in Section \ref{sec:behavior_design}: We want to develop further specialized language features to utilize motions plans and detection models within the control networks. For the concatenation of behaviors, an efficient way to express the pattern recognition within the component network's data needs to be realized. The control networks described using the eDSL presented in Section \ref{sec:control_system} operate in the reference frames of the respective virtual control systems or on joint level (cf. \emph{CTRL} in Fig. \ref{fig:basic_processing}). In order to resolve the necessary transformation between different reference frames, we need to further develop and utilize conventions for geometrical description of actions and robot parts.

	\section*{ACKNOWLEDGMENT}
	This work was supported through a grant of the German Federal Ministry of Economics and Technology (BMWi, FKZ 50 RA 1216 and FKZ 50 RA 1215).

	\bibliographystyle{abbrv}
	
	\bibliography{bibliography}

\end{document}